%% file: main.tex
\documentclass[conference]{IEEEtran}
\IEEEoverridecommandlockouts
\usepackage[numbers]{natbib}
\usepackage{amsmath,amssymb,amsfonts}
\usepackage{algorithmic}
\usepackage{graphicx}
\usepackage{textcomp}
\usepackage{float}

\usepackage{subcaption}
\usepackage[bookmarks=false]{hyperref}
\hypersetup{nolinks=true}
\usepackage{booktabs}
\usepackage{tabularx}
\newcommand{\sysml}[1]{\texttt{#1}}
\newcommand{\stereotyp}[1]{\texttt{\textless\textless PDDL\_#1\textgreater\textgreater}}
\newcommand{\element}[1]{\textit{#1}}

\def\BibTeX{{\rm B\kern-.05em{\sc i\kern-.025em b}\kern-.08em
    T\kern-.1667em\lower.7ex\hbox{E}\kern-.125emX}}

\input{listings-setup}
\input{tikz-setup}
\input{acronym-setup}

\begin{document}
\bstctlcite{IEEEexample:BSTcontrol} 

\title{Integrating AI Planning Semantics into \\SysML System Models for \\Automated PDDL File Generation\thanks{This research paper [project iMOD and LaiLa] is funded by dtec.bw – Digitalization and Technology Research Center of the Bundeswehr. dtec.bw is funded by the European Union – NextGenerationEU.}}

\author{
\IEEEauthorblockN{
Hamied Nabizada,
Tom Jeleniewski,
Lasse Beers,\\
Maximilian Weigand,
Felix Gehlhoff}

\IEEEauthorblockA{ Institute of Automation Technology\\
Helmut Schmidt University, Hamburg, Germany\\
\{firstName.lastName\}@hsu-hh.de\\}
\and 
\IEEEauthorblockN{Alexander Fay}
\IEEEauthorblockA{Chair of Automation\\
Ruhr University Bochum,
Bochum, Germany \\
alexander.fay@rub.de}
}

\maketitle
\begin{abstract}
This paper presents a SysML profile that enables the direct integration of planning semantics based on the Planning Domain Definition Language (PDDL) into system models. 
Reusable stereotypes are defined for key PDDL concepts such as types, predicates, functions and actions, while formal OCL constraints ensure syntactic consistency. 
The profile was derived from the Backus-Naur Form (BNF) definition of PDDL 3.1 to align with SysML modeling practices. 
A case study from aircraft manufacturing demonstrates the application of the profile: a robotic system with interchangeable end effectors is modeled and enriched to generate both domain and problem descriptions in PDDL format. 
These are used as input to a PDDL solver to derive optimized execution plans. 
The approach supports automated and model-based generation of planning descriptions and provides a reusable bridge between system modeling and AI planning in engineering design.
\end{abstract}


\section{Introduction}
\ac{mbse} is increasingly regarded as a viable alternative to traditional, document-based engineering approaches by enabling the continuous and structured description of production systems through formal models~\cite{Henderson.2021}. 
These models provide a unified basis for system representation, which enhances consistency and facilitates interdisciplinary collaboration throughout development processes~\cite{Schmidt.2020}.

Methods of \ac{mbse}, such as the \textit{Systems Modeling Language}~(\acs{sysml})-based frameworks described in~\cite{weilkiens2016sysmod} and~\cite{Pohl2012SPES}, provide structured means to develop consistent system models. 
They help to reduce heterogeneity in both the content and representation of subsystem models, thereby supporting reusability and comparability~\cite{Estefan.2007}. 
However, these methods often involve numerous manual steps, such as assigning process steps to appropriate technical resources. 
This task requires considerable expert knowledge and limits scalability, particularly in complex or frequently changing system configurations~\cite{Henderson.2021}. 

This becomes particularly relevant when different configurations must be evaluated to determine the optimal setup for specific optimization objectives, such as cycle time, resource utilization, or energy consumption~\cite{dahmen2018experimentable}. 
Such evaluations require analyzing both the overall system architecture and the dynamic behavior of individual subsystems~\cite{Schummer2022}. 
These aspects must be considered in relation to one another to assess the impact of structural changes on the system’s behavior~\cite{dahmen2018experimentable}.

For instance, a configuration of a robotic system with multiple end-effectors operating in parallel can achieve shorter cycle times than a configuration with only one end-effector. 
However, if the tasks are not evenly distributed or system bottlenecks occur, the additional end-effectors may experience idle periods.
In scenarios where the objective is to maximize resource utilization, such idle times can conflict with the intended optimization goal. 

In system development, it is necessary to determine which system configuration is best suited to efficiently execute a predefined set of processes while taking into account specific process constraints~\cite{Törmanen.2017}. 
Cycle time, among other metrics, can serve as a key indicator in this evaluation. 
To calculate such metrics, a process plan is required that defines both the execution order of actions and their assignment to the corresponding subsystems. 

Automation of this planning process is desirable to reduce the time required for evaluating different system configurations. 
To enable such automation, it is necessary to integrate specific planning aspects into system modeling. 
These planning aspects refer to the definition of actions needed to achieve particular goals within the modeled system, such as transporting components or assembling parts. 
They form the foundation of the research field of \textit{AI Planning}, which focuses on the development of languages and algorithms for solving planning problems~\cite{Ghallab.2016}. 

The \ac{pddl} has emerged from this field and is recognized as the de facto standard for describing planning problems~\cite{MayrDorn.2022}. 
However, the manual specification of \ac{pddl} files is acknowledged to be error-prone and time-intensive~\cite{Lindsay.2023, Sleath2024}. 
Even small mistakes can be difficult to detect and fix, as highlighted in~\cite{Strobel2020MyPDDL}. 
To address this, the generation of \ac{pddl} descriptions should be automated wherever possible, a need that has also been identified in recent research agendas for AI planning in flexible production environments~\cite{Kocher.2022}. 

This paper introduces a dedicated \acs{sysml} profile tailored for \ac{pddl} to support its use within \acs{mbse}. 
By embedding \ac{pddl}-specific constructs directly into \acs{sysml} models, the profile enables the automated and consistent generation of planning files and facilitates process planning in system development. 
The core objective of this work is to integrate symbolic planning semantics from \ac{pddl} into \acs{sysml} models in a way that enables consistent and tool-supported generation of formal planning descriptions.

The remainder of this paper is structured as follows. 
Section~\ref{sec:Fundamentals} introduces the technical foundations of \acs{sysml} and \ac{pddl} and provides essential background information. 
Section~\ref{sec:SotA} reviews related work and positions the proposed approach within the current state of research. 
Section~\ref{sec:Profile} details the developed \acs{sysml} profile for integrating \ac{pddl} planning semantics. 
Section~\ref{sec:CaseStudy} demonstrates the application of the profile in a case study from aircraft structural assembly. 
Finally, Section~\ref{sec:Conclusion} summarizes the results and outlines directions for future work.

\section{Fundamentals}
\label{sec:Fundamentals}

\subsection{SysML and Profile Mechanism}
The \ac{uml} is a standardized graphical modeling language developed by the \ac{omg}, originally intended for the design and documentation of software systems. 
It provides a wide range of elements and diagram types, such as class diagrams, activity diagrams, and sequence diagrams to represent the structure, behavior, and interactions of a system~\cite{OMGUML}. 
Although its origins lie in software engineering, \ac{uml} has also found applications in other engineering domains due to its extensibility~\cite{Stutz2002}. 

To support domain-specific modeling, the \ac{uml} metamodel includes a profile mechanism that allows for language-native extensions without altering the core specification~\cite{Weilkiens.2008}. 
These extensions introduce stereotypes, tagged values, derived properties, and constraints to tailor the language to specific needs.

A widely known and accepted extension of \ac{uml} is the \acs{sysml}, which was developed specifically for modeling complex, interdisciplinary systems across various engineering domains~\cite{ObjectManagementGroup.2019}. 
In contrast to software-centric \ac{uml}, \acs{sysml} adds capabilities for modeling requirements, physical structures, parametric constraints, and use cases~\cite{Ma.2022}. 
This makes it particularly well suited for systems engineering applications in the context of model-based approaches~\cite{Friedenthal.2014}.

While \acs{sysml} serves as a general-purpose modeling language for systems engineering, its underlying profile mechanism offers a high degree of flexibility. 
It allows for the definition of custom profiles to extend the language toward specific analysis or domain-specific objectives~\cite{Schummer2022}. 
Such profiles define reusable modeling constructs that can be systematically embedded into existing system models~\cite{seidl2012uml}.

\subsection{Planning Domain Definition Language (PDDL)} 
\ac{pddl} is a standardized language developed for describing planning problems in symbolic artificial intelligence~\cite{Ghallab.2016}. 
It distinguishes between \textit{domains} and \textit{problems}~\cite{Haslum.2019}: 
A \textit{domain} defines the structural framework by specifying general elements such as types, predicates, and actions that apply across multiple planning problems. 
A \textit{problem}, in contrast, describes a specific planning instance within a domain by specifying the available objects, the initial state, and the goal to be achieved. 
This separation enables \ac{pddl} solvers to combine one domain with multiple problem instances. 

The core elements of a domain description include \textit{types}, \textit{predicates}, \textit{actions}, and optional \textit{functions}. 
\textit{Types} are used to classify objects and constrain which objects can be associated with specific actions. 
For example, in a manufacturing domain, object types might include machines, tools, or components. 

\textit{Actions} represent operations that can be performed in the domain. 
Each action defines parameters to denote the involved objects, preconditions that must be satisfied before execution, and effects that describe how the system state changes by the execution of the action. 
These elements allow solvers to compute sequences of actions, referred to as \textit{plans}, that transform the initial state into a goal state. 
To illustrate the structure of a PDDL action, Listing~\ref{lst:assemble} shows a simplified example from a production domain. 
The action \textit{assemble-part} requires a tool to be available and results in a part being marked as assembled.

\begin{lstlisting}[language=PDDL, 
backgroundcolor = \color{light-gray},
  float=!htb,
caption={Example PDDL action: \element{assemble-part}}, 
label={lst:assemble},
basicstyle=\ttfamily\small]
(:action assemble-part
 :parameters (?p - part ?t - tool)
 :precondition (available ?t)
 :effect (assembled ?p))
\end{lstlisting}

\textit{Predicates} describe logical conditions or system states and are defined globally within the domain. 
They are referenced in the preconditions and effects of actions to express whether certain relationships between objects or properties of the system currently hold. 
For example, the predicates \textit{(available~?t)} and \textit{(assembled~?p)} in Listing~\ref{lst:assemble} indicate the availability of a tool and the assembled state of a part. 
Predicates are essential for enabling \ac{pddl} solver to determine the applicability of actions in a given state. 

\textit{Functions} represent numerical quantities such as time, cost, or distance. 
They can be used to formulate quantitative preconditions and effects, which are particularly important in optimization scenarios. 
For example, \ac{pddl} solvers can be instructed to minimize total processing time or resource consumption based on functional values.

These planning constructs are essential for enabling AI Planning tools to generate action sequences automatically based on formal domain and problem descriptions.
A comprehensive overview of the syntax and semantics of \ac{pddl} is provided in~\cite{Haslum.2019}. 

\section{Related Work}
\label{sec:SotA}

Various approaches have been proposed to bridge system modeling with logic for simulation or planning purposes. 
These methods differ in their degree of automation, the formalism used, and their integration into modeling environments such as \acs{sysml}.

In~\cite{Huckaby.2013}, a \acs{sysml}-based taxonomy for assembly tasks is presented to describe system capabilities.
These capabilities form the foundation for manually deriving \ac{pddl} actions that represent the behavior of a robotic system.
However, the approach is limited to the predefined taxonomy and does not support automated generation of PDDL Files.

Vieira et al.~\cite{vieira2023transformation} propose a capability-based approach for generating \ac{pddl} descriptions.
Using an ontology, required manufacturing capabilities are matched with the available capabilities of production resources.
This enables automatic generation of planning descriptions, but the method relies on a specific capability model and is not embedded in an \ac{mbse} context.

Rimani et al.~\cite{Rimani.2021} present a conceptual approach to modeling in \ac{hddl}, a hierarchical extension of \ac{pddl}.
The authors compare elements of \acs{sysml} and \ac{hddl} and outline a modeling workflow.
However, no transformation mechanism is provided, and the derivation of planning elements remains manual.

Batarseh and McGinnis~\cite{batarseh2012system} use a \acs{sysml} profile to automatically generate simulation models from system models.
While their goal is to reduce modeling effort and improve simulation efficiency, their approach remains limited to performance simulation and does not incorporate symbolic planning logic or support the generation of formal planning descriptions, such as \acs{pddl}. 

Another transformation-based method is presented by Wally et al.~\cite{wally2019flexible}, who derive \ac{pddl} descriptions from ISA-95-compliant models of manufacturing systems.
This method allows automatic planning based on standard-compliant representations.
However, its applicability is restricted to ISA-95 and does not generalize to \acs{sysml}-based modeling environments. 

While some existing methods offer partial automation of planning tasks, they typically lack native integration with model-based environments and do not embed planning semantics directly into system models. 

To address this gap, our previous work~\cite{Nabizada2024b} introduced a methodical approach for integrating planning capabilities into system modeling. 
The proposed workflow defines four phases, ranging from model analysis to the automated generation of \ac{pddl} descriptions, and ensures consistency between system architecture and planning logic.
The overall workflow of this approach is illustrated in \autoref{fig:workflow_model}. 

\input{img/workflow_model.tex}

The present paper contributes to Phase~II of this workflow~(highlighted in orange), which focuses on extending existing \acs{sysml} system models with planning semantics. 
To support this integration, a dedicated \acs{sysml} profile is introduced. 
This profile enables \acs{pddl} constructs such as types, predicates, functions, and actions to be directly represented within \acs{sysml} system models.

\section{Extending SysML with a Profile for PDDL Integration}
\label{sec:Profile}

To make system models usable as input for AI Planning tools, the planning semantics of \acs{pddl} must be embedded directly within the model. 
This is achieved through a dedicated \acs{sysml} profile that enables structured and tool-supported representation of planning elements. 
To ensure syntactic and semantic alignment, the profile presented here is based on the \textit{Backus-Naur Form} (BNF) specification of \ac{pddl}~3.1~\cite{Kovacs.2011}. 

An excerpt from the BNF specification of PDDL that defines the structure of an action is shown in Listing~\ref{lst:bnf_action}. 

\begin{table}[H]
\begin{lstlisting}[language={BNF}, 
                   caption={BNF definition of a PDDL action \cite{Kovacs.2011}}, 
                   label={lst:bnf_action}, 
                   backgroundcolor=\color{light-gray},
                   breaklines=false]
<action-def> ::= 
    (:action <action-symbol>
     :parameters (<typed list (variable)>)
     <action-def body>)
<action-def body> ::= 
    [:precondition <emptyOr (pre-GD)>]
    [:effect <emptyOr (effect)>]
\end{lstlisting}
\end{table}

This grammar rule specifies that every action must have a name, a parameter list, and optionally preconditions and effects. It reflects the core structure of planning logic in \ac{pddl} domains.

However, as described in~\cite{Bergmayr.2013}, direct transformation of the BNF into a modeling language metamodel is not trivial, as it fails to account for structural and semantic differences between textual and graphical representations.
For this reason, the profile was developed manually, following the \ac{pddl} specification and adapting its constructs to the \acs{sysml} modeling context.

As part of Phase~II of the workflow illustrated in \autoref{fig:workflow_model}, this paper now details the structure of the developed \acs{sysml} profile.\footnote{The profile is available on GitHub: \url{https://github.com/hsu-aut/SysML-Profile-PDDL}}   
The profile enables the annotation of system model elements with \ac{pddl} constructs such as types, predicates, functions, and actions.  
These enriched system models provide the foundation for Phase~IV of the workflow, which automates the generation of \ac{pddl} domain files, as supported by the transformation algorithm introduced in our earlier work~\cite{Nabizada2024c}. 

A system model typically provides the structural and behavioral information required to define a \ac{pddl} domain. 
The corresponding problem descriptions, which include initial states and goals, are often derived from product-specific data and are not considered in this paper. 
The focus is placed on the domain description, which can be specified using modeling elements commonly found in system models~\cite{Nabizada2024b}. 

Each \ac{pddl} construct is represented as a stereotype in the profile. 
These stereotypes extend existing \ac{uml}/\acs{sysml} metaclasses and introduce additional tagged values and constraints where necessary. 
\autoref{tab:pddl_to_sysml} summarizes how each core \ac{pddl} construct is mapped to a specific \acs{sysml} stereotype and its corresponding metaclass.  
This structural mapping helps ensure that planning semantics are embedded in a way that is both tool-compatible and intuitive for system engineers. 

\begin{table}[h]
\centering
\caption{Mapping of PDDL Concepts to SysML Elements}
\label{tab:pddl_to_sysml}
\begin{tabular}{>{\raggedright\arraybackslash}p{2.32cm} >{\raggedright\arraybackslash}p{2.7cm} >{\raggedright\arraybackslash}p{2.4cm}}
\toprule
\textbf{PDDL Concept} & \textbf{Applied Stereotype} & \textbf{Extended SysML Metaclass} \\
\midrule
Domain Definition & \texttt{<<PDDL\_Domain>>} & \texttt{Model}, \newline \texttt{Package} \\
Object Type & \texttt{<<PDDL\_Type>>} & \texttt{Class} \\
Logical Condition & \texttt{<<PDDL\_Predicate>>} & \texttt{ObjectFlow}, \newline \texttt{ControlFlow} \\
Numeric Expression & \texttt{<<PDDL\_Function>>} & \texttt{ObjectFlow}, \newline \texttt{ControlFlow} \\
Action Specification & \texttt{<<PDDL\_Action>>} & \texttt{CallBehavior\newline Action} \\
\bottomrule
\end{tabular}
\end{table}

The stereotype \stereotyp{Domain} designates the top-level context for planning within the system model and extends \sysml{Model} or \sysml{Package}. 
This makes it possible to define a complete planning domain within a coherent modeling container. 
\stereotyp{Type} extends \sysml{Class} and enables the modeling of object types and hierarchies, as required in \ac{pddl} to define parameter constraints and inheritance.

\stereotyp{Action} represents operations within the domain and extends \sysml{CallBehaviorAction}, commonly used in activity diagrams to model system behavior~\cite{beers2024sysml}. 
Preconditions and effects are modeled through derived properties referencing incoming and outgoing control or object flows, which are stereotyped as \stereotyp{Predicate} or \stereotyp{Function}.

The stereotype \stereotyp{Predicate} allows modeling logical conditions, while \stereotyp{Function} captures numerical relationships, e.g., cost or distance metrics.
Both extend the metaclasses \sysml{ObjectFlow} and \sysml{ControlFlow}, thereby supporting their use in behavioral diagrams.
\sysml{Customization} elements are used to add derived properties and constraints and to restrict stereotype usage to specific model elements. 

To complement the structural mapping, the profile integrates validation mechanisms to enforce modeling correctness. 
The \textit{Object Constraint Language} (OCL) is used to define formal constraints that guide modelers and tools in maintaining syntactic compliance with the \ac{pddl} specification. 
Such constraints validate element properties at modeling time and help detect specification errors early in the modeling process. 

For instance, \autoref{lst:ocl_domainname} shows a constraint applied to \stereotyp{Domain} elements, which ensures that each domain name is defined and conforms to the identifier rules of \ac{pddl}. 

\begin{table}[H]
\begin{lstlisting}[language=OCL, caption={OCL constraint for validating the identifier of a PDDL domain}, label={lst:ocl_domainname}, backgroundcolor = \color{light-gray}]
context PDDL_Domain inv ValidateDomainName:
  not self.name.oclIsUndefined() and
  self.name.matches('^[a-zA-Z][a-zA-Z0-9_]*$')
\end{lstlisting}
\end{table}

This constraint enforces that domain identifiers begin with a letter and may be followed by alphanumeric characters or underscores, thus aligning with the syntax defined in the \ac{pddl} standard.
Further OCL constraints can be applied to other stereotypes, such as validating the completeness of \stereotyp{Action} definitions or checking for consistent use of predicates across actions. 
Such validation constraints go beyond syntactic validation and help ensure that the structure of the system model aligns with the semantic requirements of \ac{pddl}. 
Structural support of this kind is particularly relevant in complex domains, where inconsistencies and redundancies can arise even in carefully designed models~\cite{Shah2013}. 

For example, one important aspect of structural consistency is the uniqueness of type definitions. 
In \ac{pddl}, each type identifier within a domain must be unique. 
Duplicate type names may lead to ambiguities during domain interpretation or cause errors in the planning toolchain~\cite{Sleath2024}. 
To prevent such conflicts at modeling time, the profile includes a constraint that enforces unique names across all \stereotyp{Type} elements within a domain. 

\autoref{lst:ocl_typenames} shows the OCL constraint applied to model elements stereotyped as \stereotyp{Domain}. 
It filters the set of all owned elements to extract those that are instances of \stereotyp{Type} and checks whether each of them has a distinct \texttt{name} property.

Here, the expression \texttt{select(...)} is used to filter elements of type \stereotyp{Type}, while \texttt{isUnique(...)} ensures that no two such elements share the same name. 
By defining such constraints at the metamodel level, the profile helps to shift error detection from the code generation or execution phase to the modeling phase, thereby reducing the time and effort required for debugging and validation.

\begin{table}[h]
\begin{lstlisting}[language=OCL, caption={OCL constraint for enforcing unique type names within a domain}, label={lst:ocl_typenames}, backgroundcolor = \color{light-gray}] 
context PDDL_Domain inv UniqueTypeNames: 
  self.ownedElement 
    ->select(e | e.oclIsKindOf(PDDL_Type)) 
    ->isUnique(t | t.name) 
\end{lstlisting}
\end{table}

To clarify the steps of Phase~II in Figure~\ref{fig:workflow_model}, the previously introduced \textit{assemble-part} action from Listing~\ref{lst:assemble} serves as an illustrative example. 
In a typical system model, elements such as \textit{tool} and \textit{part} are already defined as system types, and behavioral steps like \textit{assemble-part} are modeled as actions with incoming and outgoing flows. 

During Phase~II, these existing elements are annotated with stereotypes from the profile: \stereotyp{Type} for system types, \stereotyp{Predicate} for state-relevant flows such as \textit{available} or \textit{assembled}, and \stereotyp{Action} for behavioral nodes. 
To ensure modeling correctness, OCL constraints are applied. 
For instance, the type names \textit{tool} and \textit{part} must be unique within the annotated domain model, as enforced by the constraint in Listing~\ref{lst:ocl_typenames}.

By annotating system models using this profile, a consistent representation of domain-specific planning constructs is achieved that aligns with both \acs{sysml} semantics and \ac{pddl} requirements. 
The annotated elements serve as the input for an algorithm that extracts relevant model content and automatically generates a syntactically valid \ac{pddl} domain file, as detailed in~\cite{Nabizada2024c}. 
While the primary focus of this paper lies on domain generation, the enriched model also forms the structural foundation for generating problem descriptions. 
These rely on the same planning constructs and require their manual association with instance-specific data to ensure semantic consistency. 
Both files are generated automatically using template logic that extracts and formats the required content, resulting in a complete and tool-supported specification of planning problems.

\section{Case Study in Aircraft Manufacturing}
\label{sec:CaseStudy}
To validate the developed profile, a case study from structural assembly in aircraft manufacturing was conducted. 
The case study is part of the \textit{iMOD} project, which aims to address key challenges in flexible and efficient aircraft production, including the integration of robotics and system design~\cite{Gehlhoff2022iMOD}. 

\begin{figure}[h]
    \centering
    \includegraphics[width=0.8\linewidth]{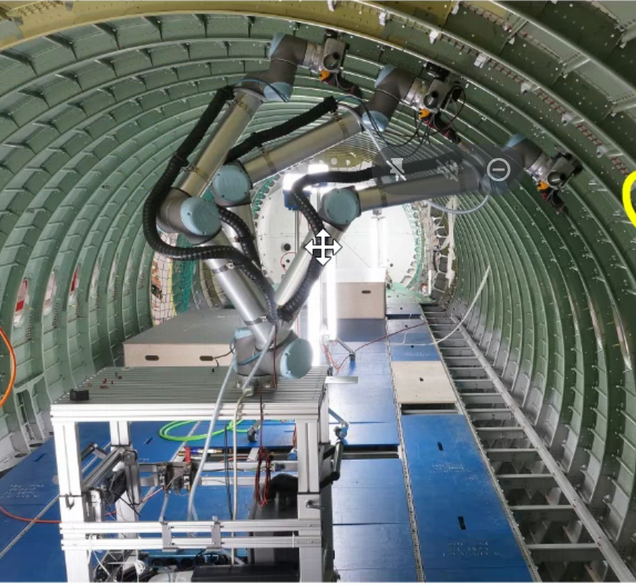}
    \caption{\textit{Collar Screwing System} inside the fuselage \cite{beers2023mbse}}
    \label{fig:css}
\end{figure}

A movable UR10 robotic arm is positioned within the aircraft fuselage and used to screw various collars onto rivets, with each collar type requiring a specific end effector. 
This system (referred to as \textit{Collar Screwing System}) is illustrated in \autoref{fig:css}. 
The approach for creating the corresponding system model is described in~\cite{beers2023mbse}.

\begin{figure*}[htp]
    \centering
    \includegraphics[width=1\textwidth]{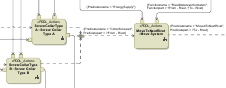}
    \caption{Excerpt from the enriched system model of the \textit{Collar Screwing System}}
    \label{fig:systemodell}
\end{figure*}

In this case study, the planning objective was to identify a process plan that minimizes cycle time while accounting for hardware reconfiguration, which is necessary when switching between different end effectors to handle the correct rivet type. 
An excerpt from the activity diagram of the enriched system model illustrating the interaction between modeled system behavior and the \ac{pddl} profile is shown in \autoref{fig:systemodell}.

The model excerpt shows three \stereotyp{Action} elements, each with associated preconditions and effects. 
The \stereotyp{Action} elements \element{ScrewCollarType A} and \element{ScrewCollarType B} describe the process steps for screwing the respective collar type. 
Once this process step is completed, the system moves to the next rivet in the \stereotyp{Action} \element{MoveToNextRivet}. 
These actions are executed repeatedly during operation to process multiple rivets. 

To execute the action \element{MoveToNextRivet}, the following preconditions must be fulfilled:
\begin{enumerate}
    \item The system must be supplied with energy. Thus, the predicate \element{EnergySupply} flows into the action.
    \item The collar currently engaged by the end effector must already be screwed. The actions \element{ScrewCollarTypeA} and \element{ScrewCollarTypeB} are responsible for setting the predicate \element{CollarScrewed} to \texttt{True} once the action is completed.
    \item The coordinates of the rivets and the associated distances between them are defined in the \sysml{Function} \element{RivetDistanceInformation} and are passed from there to the \stereotyp{Action} \element{MoveToNextRivet}. A \ac{pddl} solver can identify the next rivet to be processed based on the distances and rivet types. 
\end{enumerate}

Once the action is executed, its effect is that the end effector is positioned at a new working location.
The corresponding predicate \element{MovedToNextRivet} is set to \texttt{True} with the new rivet position.

In the case study, \textit{Magic Systems of Systems Architect}~(\acs{msosa}) was used as the modeling environment for creating both the system model and the applied profile. 
The system model was then enriched with planning semantics through stereotype annotations as defined by the profile. 
The enriched system model can systematically be queried for planning information. 
To enable automated generation of planning descriptions, the algorithm described in~\cite{Nabizada2024c} was used. 
This approach leverages the \textit{Report Wizard} functionality in \acs{msosa}, which utilizes the \textit{Velocity} engine and template logic defined in the \ac{vtl}. 
Domain- and problem-specific templates were created, each containing the syntactic structure of \ac{pddl} and placeholder logic to extract the required model content. 

To demonstrate the result of the automated transformation, a section of the generated \ac{pddl} domain description is provided in \autoref{lst:pddl}. 
The action \element{MoveToNextRivet}, as modeled in the system diagram shown in \autoref{fig:systemodell}, serves as the basis for this example. 
Using the algorithm described in~\cite{Nabizada2024c}, the annotated model elements were queried and interpreted according to their semantics and associated stereotype properties. 
The generated \ac{pddl} action comprises three components: a set of parameters that define the object types involved in the action, a list of preconditions that must be satisfied for execution, and a list of effects describing the resulting changes to the system state. 
Each of these elements was extracted from the model by querying stereotype attributes and derived properties, such as the predicates and functions assigned to incoming or outgoing flows. 

In this specific case, the predicate \element{CollarScrewed} is referenced as a precondition to ensure that the previous assembly step has been completed before the robot moves.
Additionally, \element{EnergySupply} is included as a logical requirement for executing the action.
The function \element{RivetDistanceInformation} is used to calculate the distance-based cost of the movement, which is expressed using the \ac{pddl} \texttt{:increase} construct.

{\footnotesize
\begin{lstlisting}[
backgroundcolor = \color{light-gray},
  float=!htb,
  caption={PDDL syntax for the action \element{MoveToNextRivet}},
  label={lst:pddl},
  language=PDDL]
(:action MoveToNextRivet
   :parameters (?From - Rivet ?To - Rivet)
   :precondition 
        (and 
            (CollarScrewed ?From)
            (EnergySupply)
        )
    :effect 
        (and  
            (MovedToNextRivet ?To ) 
            (increase 
               (total-cost) 
               (RivetDistanceInformation ?From ?To)) 
        )
)
\end{lstlisting}
}

To complete the planning input, a problem description must also be specified in \ac{pddl} format. 
While the domain file defines reusable constructs such as types, predicates, and functions, the problem file provides problem-specific information by declaring object instances, assigning initial conditions, and specifying goal criteria. 
In this case study, relevant product information such as the positions, types, and quantities of rivets was automatically extracted from the product structure in \textit{3DExperience} and imported into the \acs{msosa} modeling environment. 
The assignment of this data to the planning constructs defined at the domain level was carried out manually to ensure semantic consistency between system model and planning logic.
Based on the resulting model, a \ac{vtl} template was applied to automatically generate a valid \ac{pddl} problem description by extracting and formatting all required instance data. 

Once both domain and problem descriptions are available, they can be passed to a solver to solve the planning problem. 
For example, in this case study, the solver \textit{Delfi} \cite{Katz.2018} provided by \cite{Muise.2022} was used. 
In the evaluated scenario, two different rivet types were present within the robot’s workspace, each requiring a dedicated tool. 
The \ac{pddl} solver generated an optimized sequence of movement and assembly actions that minimized overall cost by jointly considering travel distance and the number of tool changes. 
As a result, the plan first completed all tasks for the first rivet type before performing a single reconfiguration and continuing with the remaining tasks.

With the resulting plan from the solver, the system can finally be evaluated regarding optimization goals such as minimizing cycle time or expected energy consumption. 
The resulting data serves as the basis for system engineers' decisions during system design. 
For example, if a UR16 is considered instead of a UR10, a direct comparison between these two configurations can be performed. 
For this, the system model must be expanded with information for the UR16, but the same planning logic can be reused without additional effort. 

While the case study focuses on aircraft manufacturing, the profile itself is domain-independent and applicable to other system modeling scenarios involving symbolic planning.

\section{Conclusion and Outlook} 
\label{sec:Conclusion}

This paper presents a \acs{sysml} profile that enables the integration of planning semantics from \acs{pddl} directly into system models.
The profile maps essential \acs{pddl} concepts, including domains, types, predicates, functions, and actions, to reusable modeling constructs using \acs{sysml} stereotypes.
By aligning these planning constructs with standard modeling practices, the profile supports consistent, structured, and model-based generation of domain descriptions.

The profile was applied in a case study from aircraft manufacturing, where a robotic system for collar screwing was modeled and enriched with planning annotations. 
Product-specific data, such as rivet positions and types, were imported from the product structure and used to create a problem description. 
Both domain and problem files were generated automatically using template-based reporting and served as input for a \ac{pddl} solver to evaluate system behavior under defined optimization objectives. 

The results demonstrate that planning logic can be systematically represented at the modeling level, making it easier to generate planning descriptions across different configurations.
Once embedded in the system model, planning constructs remain consistent and reusable, even as the system evolves during design.

While the profile enables a formalized representation of planning constructs, the manual annotation of system models remains a time-consuming task. 
This modeling overhead may hinder adoption in practice, especially in industrial settings where modeling resources are limited. 
To address this limitation, a profile-driven assistance system is planned that actively supports users during the enrichment of system models. 
This system will provide interactive guidance throughout the modeling process, including context-sensitive suggestions for applicable stereotypes, automatic checks for syntactic and semantic consistency, and visual warnings highlighting missing or invalid planning constructs. 
By integrating the underlying rules of the profile and linking them to the modeling environment, the assistance system aims to reduce modeling errors and accelerate the creation of valid planning annotations. 

In addition, future extensions could explore how planning-relevant instance data from external sources, such as product models or engineering databases, can be imported and mapped to the existing planning constructs. 
Building on approaches such as~\cite{Weigand.2022}, which demonstrate how to extract and structure model information from \acs{msosa}, future research could explore the integration of additional engineering data sources to further enhance the completeness and consistency of planning inputs. 
Furthermore, recent advances such as the \textit{Unified Planning Library}~\cite{Micheli2025UnifiedPlanning} provide a generic planning infrastructure that facilitates the connection between planning logic and heterogeneous data sources. 

Moreover, recent advancements in SysML v2~\cite{ObjectManagementGroup.2024} could be leveraged to improve the integration of planning semantics by utilizing new validation and metamodeling features. These features, including enhanced support for model consistency and automated validation, could further streamline the process of incorporating planning semantics into system models, offering a more robust and scalable solution for complex engineering systems. 
While this work is based on SysML v1.6, the transition to SysML v2 is being considered as tool maturity increases. 

\interlinepenalty=10000
\bibliographystyle{IEEEtran}
\bibliography{references.bib}

\end{document}

%% file: listings-setup.tex
\usepackage{listings}
\usepackage[dvipsnames]{xcolor}
\definecolor{light-gray}{gray}{0.95} 
\lstdefinelanguage{PDDL}
{
  sensitive=false,    
  morecomment=[l]{;}, 
  alsoletter={:,-},  
  morekeywords={
    define,domain,problem,not,and,or,when,forall,exists,either,
    :domain,:requirements,:types,:objects,:constants,
    :predicates,:action,:parameters,:precondition,:effect,
    :fluents,:primary-effect,:side-effect,:init,:goal,
    :strips,:adl,:equality,:typing,:conditional-effects,
    :negative-preconditions,:disjunctive-preconditions,
    :existential-preconditions,:universal-preconditions,:quantified-preconditions,
    :functions,assign,increase,decrease,scale-up,scale-down,
    :metric,minimize,maximize,
    :durative-actions,:duration-inequalities,:continuous-effects,
    :durative-action,:duration,:condition
  }
}
\lstdefinelanguage{OCL}{
    morekeywords={context, inv, self, and, not, matches},
    sensitive=true,
    basicstyle=\footnotesize\ttfamily,
    keywordstyle=\bfseries,
    commentstyle=\itshape\color{gray},
    breaklines=true,
}

\lstdefinelanguage{BNF}{
    morekeywords={::=, |},
    sensitive=false,
    alsoletter={<>=-},
    moredelim=**[is]{<}{>},
    keywordstyle=\bfseries,
    basicstyle=\ttfamily\footnotesize,
}

%% file: tikz-setup.tex
\usepackage{tikz}
\usetikzlibrary{shapes,snakes, calc, arrows.meta, positioning, fit}
\tikzset{
	>=Latex,
	line/.style={draw,->},
	anode/.style={rectangle,draw,
		align=center,rounded corners,minimum height=4em,font=\strut},
	bnode/.style={anode,fill=white, font=\strut},
	cnode/.style={anode, fill=cyan!20, font=\strut},
treenode/.style = {circle,
	draw=black,thick, fill=white, align=center, minimum size=1cm},
root/.style     = {treenode, font=\footnotesize},
env/.style      = {treenode, font=\footnotesize}, 
dummy/.style    = {circle,draw}
}
\tikzstyle{decision} = [diamond, draw, fill=yellow!20, 
    text width=6em, text badly centered, node distance=3cm, inner sep=0pt, line width=0.8pt, minimum height=9em,
    minimum width=9em]
\tikzstyle{block} = [rectangle, draw, fill=cyan!20, 
    text width=6.7em, text centered, rounded corners, minimum height=4em, line width=0.8pt]
\tikzstyle{line} = [draw, -latex, line width=0.8pt]
\tikzstyle{dashline} = [draw, -latex,dashed, line width=0.8pt]
\tikzstyle{helpline} = [draw, line width=0.8pt]
\tikzstyle{method} =  [trapezium, trapezium left angle=80, trapezium right angle=-80,text centered,text width = 2cm,minimum height=1cm, minimum width=2cm, draw=black, fill=yellow!20, line width=0.8pt]
\tikzstyle{exArtifact} =[rectangle, draw, fill=cyan!20, 
    text width=6.7em, text centered, rounded corners, minimum height=4em, line width=0.8pt]
\tikzstyle{genArtifact} =[rectangle, draw, fill=green!20, 
    text width=6.7em, text centered, rounded corners, minimum height=4em, line width=0.8pt]

%% file: acronym-setup.tex
\usepackage[nolist]{acronym}
\begin{acronym}
\acro{mbse}[MBSE]{\textit{Model-Based Systems Engineering}}
\acro{sysml}[SysML]{\textit{Systems Modeling Language}}
\acro{msosa}[MSoSA]{Magic Systems of Systems Architect}
\acro{bnf}[BNF]{\textit{Backus-Naur-Form}}
\acro{pddl}[PDDL]{\textit{Planning Domain Definition Language}}
\acro{uml}[UML]{\textit{Unified Modeling Language}}
\acro{vtl}[VTL]{\textit{Velocity Template Language}}
\acro{ocl}[OCL]{\textit{Object Constraint Language}}
\acro{omg}[OMG]{\textit{Object Management Group}}
\acro{hddl}[HDDL]{\textit{Hierarchical Domain Definition Language}}
\end{acronym}

%% file: img/workflow_model.tex
\definecolor{HighlightPhase}{RGB}{255,50,0} 
\begin{figure*}[h]
    \centering
    \resizebox{1\textwidth}{!}{
    \begin{tikzpicture}[node distance= 2.9cm]
        \node[exArtifact](sysModel){System Model};
        \node [method, right of= sysModel] (analyzeSM) {Analyze System Model};
        \node [method, right of= analyzeSM] (scope) {Define Scope of Observation};
        \node [method, right of= scope] (identRelElements) {Identify relevant Elements};

        
        \node[exArtifact, below of= sysModel](profile){PDDL Profile};
        \node[method, right of= profile](genDomain){Create PDDL Domain (excl. Actions)};
        \node[method, right of= genDomain](defAction){Define Actions of Domain};
        \node[genArtifact, right of= defAction](aufbSM){Extended System Model};
        \node[below=0.3cm of profile] {\textit{ focus of this work}};

        \node[exArtifact, below of= profile](product){Product Model};
        \node[method, right of= aufbSM](genPD){Generate PDDL Descriptions};
        \node[method, right of = product](identProdInfo){Identify Product Information};
        \node[method, right of = identProdInfo](extract){Extract Relevant Information};
        \node[method, right of = extract](transmit){Transfer to MBSE Environment};
        \node[method, right of = transmit](annot){Annotate according to PDDL Domain};
        \node[genArtifact, right of = annot](aufbPM){Extended Product Model};

        \node[exArtifact, above of= genPD](algorithm) {Algorithm};
        \node[genArtifact, right of = genPD](problem){PDDL Domain \& Problem};
        \node[method, right of = problem](solve){Solve Problem Description};
        \node[genArtifact, above of = solve](plan){PDDL Plan};

        \node[draw, dotted, line width=0.8pt,fit=(sysModel)(analyzeSM)(scope)(identRelElements), inner sep=0.1cm] (phase1) {};
        \node[rotate=90, above=0cm] at (phase1.west) {\textbf{Phase I}};


        \node[draw=HighlightPhase, ultra thick, dashed, line width=1.2pt,
  fit=(profile)(genDomain)(defAction)(aufbSM),
  inner sep=0.1cm] (phase2) {};
        \node[rotate=90, above=0cm] at (phase2.west) {\textbf{Phase II}};

        \node[draw, dotted, line width=0.8pt,fit=(product)(identProdInfo)(extract)(annot)(aufbPM), inner sep=0.1cm] (phase3) {};
        \node[rotate=90, above=0cm] at (phase3.west) {\textbf{Phase III}};

        \node[draw, dotted,line width=0.8pt, fit=(algorithm)(problem), inner sep=0.1cm] (phase4) {};
        \node[ above=0cm] at (phase4.north) {\textbf{Phase IV}};

        \path[line] (sysModel) -- (analyzeSM);
        \path[line] (analyzeSM) -- (scope);
        \path[line] (scope) -- (identRelElements);
        \path[line] (identRelElements) -- ++(0,-1.5cm) -| (genDomain);
        \path[line](profile) -- (genDomain);
        \path[line](genDomain) -- (defAction);
        \path[line](defAction) -- (aufbSM);
        \path[line](aufbSM) -- (genPD);
        \path[line](product) -- (identProdInfo);
        \path[line](identProdInfo) -- (extract);
        \path[line](extract) -- (transmit);
        \path[line](transmit) -- (annot);
        \path[line](annot) -- (aufbPM);
        \path[line](aufbPM) |- ++(0,1.7cm) -| (genPD);
        \path[dashline](genDomain) |- ++(0,-1.5cm)-| (annot);
        \path[line](algorithm)--(genPD);
        \path[line](genPD)  -- (problem);
        \path[line](problem)  -- (solve); 
        \path[line](solve)  -- (plan); 
        
    \end{tikzpicture}
    }
    \caption{Workflow model for automated generation of PDDL descriptions (adapted from~\cite{Nabizada2024b})}
    \label{fig:workflow_model}
\end{figure*}

%% file: main.bbl
\begin{thebibliography}{10}
\providecommand{\url}[1]{#1}
\csname url@samestyle\endcsname
\providecommand{\newblock}{\relax}
\providecommand{\bibinfo}[2]{#2}
\providecommand{\BIBentrySTDinterwordspacing}{\spaceskip=0pt\relax}
\providecommand{\BIBentryALTinterwordstretchfactor}{4}
\providecommand{\BIBentryALTinterwordspacing}{\spaceskip=\fontdimen2\font plus
\BIBentryALTinterwordstretchfactor\fontdimen3\font minus \fontdimen4\font\relax}
\providecommand{\BIBforeignlanguage}[2]{{%
\expandafter\ifx\csname l@#1\endcsname\relax
\typeout{** WARNING: IEEEtran.bst: No hyphenation pattern has been}%
\typeout{** loaded for the language `#1'. Using the pattern for}%
\typeout{** the default language instead.}%
\else
\language=\csname l@#1\endcsname
\fi
#2}}
\providecommand{\BIBdecl}{\relax}
\BIBdecl

\bibitem{Henderson.2021}
K.~Henderson and A.~Salado, ``{Value and benefits of model--based systems engineering (MBSE): Evidence from the literature},'' \emph{Systems Engineering}, vol.~24, no.~1, pp. 51--66, 2021.

\bibitem{Schmidt.2020}
M.~M. Schmidt and R.~Stark, ``{Model-Based Systems Engineering (MBSE) as computer-supported approach for cooperative systems development},'' in \emph{{Proceedings of 18th European Conference on Computer-Supported Cooperative Work}}, 2020.

\bibitem{weilkiens2016sysmod}
T.~Weilkiens, \emph{{SYSMOD - The Systems Modeling Toolbox: Pragmatic MBSE with SysML}}, 3rd~ed.\hskip 1em plus 0.5em minus 0.4em\relax Fredesdorf: MBSE4U, 2020.

\bibitem{Pohl2012SPES}
K.~Pohl, H.~H{\"o}nninger, R.~Achatz, and M.~Broy, Eds., \emph{{Model-Based Engineering of Embedded Systems: The SPES 2020 Methodology}}.\hskip 1em plus 0.5em minus 0.4em\relax Berlin, Heidelberg: Springer, 2012.

\bibitem{Estefan.2007}
J.~A. Estefan, ``{Survey of Model-based Systems Engineering (MBSE) Methodologies},'' \emph{Incose MBSE Focus Group}, vol.~25, no.~8, 2007.

\bibitem{dahmen2018experimentable}
U.~Dahmen and J.~Rossmann, ``{Experimentable digital twins for a modeling and simulation-based engineering approach},'' in \emph{2018 IEEE International Systems Engineering Symposium (ISSE)}, 2018.

\bibitem{Schummer2022}
S.~Schummer and P.~Hyba, ``{An approach for system analysis with model-based systems engineering and graph data engineering},'' \emph{Data-Centric Engineering}, vol.~3, 2022.

\bibitem{Törmanen.2017}
M.~Törmänen, A.~Hägglund, T.~Rocha, and E.~Drenth, ``{Integrating Multi-Disciplinary Optimization into the Product Development Process using Model-Based Systems Engineering (MBSE)},'' in \emph{NAFEMS World Congress}, 2017.

\bibitem{Ghallab.2016}
M.~Ghallab, D.~Nau, and P.~Traverso, \emph{{Automated Planning and Acting}}.\hskip 1em plus 0.5em minus 0.4em\relax Cambridge: {Cambridge University Press}, 2016.

\bibitem{MayrDorn.2022}
C.~Mayr-Dorn, A.~Egyed, M.~Winterer, C.~Salomon, and H.~F{\"u}rschu{\ss}, ``{Evaluating PDDL for programming production cells: a case study},'' in \emph{IEEE/ACM 4th International Workshop on Robotics Software Engineering (RoSE)}, 2022.

\bibitem{Lindsay.2023}
A.~Lindsay, ``{On Using Action Inheritance and Modularity in PDDL Domain Modelling},'' \emph{Proceedings of the International Conference on Automated Planning and Scheduling}, vol.~33, no.~1, 2023.

\bibitem{Sleath2024}
K.~Sleath and P.~Bercher, ``{Detecting {AI} Planning Modelling Mistakes -- Potential Errors and Benchmark Domains},'' in \emph{PRICAI 2023: Trends in Artificial Intelligence}, ser. Lecture Notes in Computer Science, F.~Liu, L.~Shen, H.~C. Lau, P.~Lukowicz, F.~Chen, H.~Yan, and B.~An, Eds., vol. 14326.\hskip 1em plus 0.5em minus 0.4em\relax Springer, 2024.

\bibitem{Strobel2020MyPDDL}
V.~Strobel and A.~Kirsch, ``{MyPDDL: Tools for Efficiently Creating PDDL Domains and Problems},'' in \emph{{Knowledge Engineering Tools and Techniques for AI Planning}}, M.~Vallati and D.~Kitchin, Eds.\hskip 1em plus 0.5em minus 0.4em\relax Cham: Springer International Publishing, 2020.

\bibitem{Kocher.2022}
A.~K{\"o}cher, R.~Heesch, N.~Widulle, A.~Nordhausen, J.~Putzke, A.~Windmann, and O.~Niggemann, ``{A Research Agenda for AI Planning in the Field of Flexible Production Systems},'' in \emph{2022 IEEE 5th International Conference on Industrial Cyber-Physical Systems (ICPS)}, 2022.

\bibitem{OMGUML}
\BIBentryALTinterwordspacing
{Object Management Group}, ``{\textit{Unified Modeling Language (UML), Version 2.5.1}},'' 2017. [Online]. Available: \url{https://www.omg.org/spec/UML/}
\BIBentrySTDinterwordspacing

\bibitem{Stutz2002}
C.~Stutz, J.~Siedersleben, D.~Kretschmer, and W.~Krug, ``{Analysis beyond UML},'' in \emph{Proceedings of the 10th IEEE International Requirements Engineering Conference (RE 2002)}, 2002.

\bibitem{Weilkiens.2008}
T.~Weilkiens, \emph{{Systems engineering with SysML/UML: Modeling, analysis, design}}.\hskip 1em plus 0.5em minus 0.4em\relax Amsterdam and Boston: {Morgan Kaufmann/Elsevier}, 2008.

\bibitem{ObjectManagementGroup.2019}
\BIBentryALTinterwordspacing
{Object Management Group}, ``{\textit{Systems Modeling Language (SysML), Version 1.6}},'' 2019. [Online]. Available: \url{https://www.omg.org/spec/SysML/1.6/}
\BIBentrySTDinterwordspacing

\bibitem{Ma.2022}
J.~Ma, G.~Wang, J.~Lu, H.~Vangheluwe, D.~Kiritsis, and Y.~Yan, ``{Systematic Literature Review of MBSE Tool-Chains},'' \emph{Applied Sciences}, vol.~12, no.~7, 2022.

\bibitem{Friedenthal.2014}
S.~Friedenthal, A.~Moore, and R.~Steiner, \emph{{A Practical Guide to SysML: The Systems Modeling Language}}, third edition~ed.\hskip 1em plus 0.5em minus 0.4em\relax Waltham, USA: {Morgan Kaufmann}, 2014.

\bibitem{seidl2012uml}
M.~Seidl, M.~Scholz, C.~Huemer, and G.~Kappel, \emph{{UML @ Classroom: An Introduction to Object-Oriented Modeling}}.\hskip 1em plus 0.5em minus 0.4em\relax Cham: Springer, 2015.

\bibitem{Haslum.2019}
P.~Haslum, N.~Lipovetzky, D.~Magazzeni, and C.~Muise, \emph{{An Introduction to the Planning Domain Definition Language}}.\hskip 1em plus 0.5em minus 0.4em\relax Cham: {Springer International Publishing}, 2019.

\bibitem{Huckaby.2013}
J.~Huckaby, S.~Vassos, and H.~I. Christensen, ``{Planning with a task modeling framework in manufacturing robotics},'' in \emph{2013 IEEE/RSJ International Conference on Intelligent Robots and Systems}, 2013.

\bibitem{vieira2023transformation}
L.~M. Vieira~da Silva, R.~Heesch, A.~K{\"o}cher, and A.~Fay, ``{Transformation eines F{\"a}higkeitsmodells in einen PDDL-Planungsansatz},'' \emph{at-Automatisierungstechnik}, vol.~71, no.~2, 2023.

\bibitem{Rimani.2021}
J.~Rimani, C.~Lesire, S.~Lizy-Destrez, and N.~Viola, ``{Application of MBSE to model Hierarchical AI Planning problems in HDDL},'' in \emph{International Conference on Automated Planning and Scheduling (ICAPS)}, 2021.

\bibitem{batarseh2012system}
O.~Batarseh and L.~F. McGinnis, ``{System modeling in SysML and system analysis in Arena},'' in \emph{Proceedings of the Winter Simulation Conference (WSC)}, 2012.

\bibitem{wally2019flexible}
B.~Wally, J.~Vysko{\v{c}}il, P.~Nov{\'a}k, C.~Huemer, R.~{\v{S}}indel{\'a}r, P.~Kadera, A.~Mazak, and M.~Wimmer, ``{Flexible production systems: Automated generation of operations plans based on ISA-95 and PDDL},'' \emph{IEEE Robotics and Automation Letters}, vol.~4, no.~4, 2019.

\bibitem{Nabizada2024b}
H.~Nabizada, T.~Jeleniewski, F.~Gehlhoff, and A.~Fay, ``{Model-Based Workflow for the Automated Generation of PDDL Descriptions},'' in \emph{Proceedings of the 29th IEEE International Conference on Emerging Technologies and Factory Automation (ETFA)}, 2024.

\bibitem{Kovacs.2011}
D.~L. Kovacs, ``{Complete BNF description of PDDL 3.1},'' \emph{Language Specification, Department of Measurement and Information Systems, Budapest University of Technology and Economics}, 2011.

\bibitem{Bergmayr.2013}
A.~Bergmayr and M.~Wimmer, ``{Generating Metamodels from Grammars by Chaining Translational and By-Example Techniques},'' in \emph{MDEBE@ MoDELS}, 2013.

\bibitem{Nabizada2024c}
H.~Nabizada, T.~Jeleniewski, L.~Beers, F.~Gehlhoff, and A.~Fay, ``{Automated PDDL Domain File Generation for Enhancing Production System Development based on SysML Models},'' in \emph{Proceedings of the AI4CC-IPS-RCRA-SPIRIT 2024}, November 2024.

\bibitem{beers2024sysml}
L.~Beers, H.~Nabizada, M.~Weigand, F.~Gehlhoff, and A.~Fay, ``{A SysML Profile for the Standardized Description of Processes during System Development},'' in \emph{2024 IEEE International Systems Conference (SysCon)}, 2024.

\bibitem{Shah2013}
M.~Shah, L.~Chrpa, D.~Kitchin, T.~McCluskey, and M.~Vallati, ``{Exploring knowledge engineering strategies in designing and modelling a road traffic accident management domain},'' in \emph{2013 Proceedings of the 23rd International Joint Conference on Artificial Intelligence (IJCAI)}, 2013.

\bibitem{Gehlhoff2022iMOD}
F.~Gehlhoff, H.~Nabizada, M.~Weigand, L.~Beers, O.~Ismail, A.~Wenzel, A.~Fay, P.~Nyhuis, W.~Lagutin, and M.~Röhrig, ``{Challenges in Automated Commercial Aircraft Production},'' in \emph{IFAC PapersOnLine}, vol.~55, no.~2, 2022.

\bibitem{beers2023mbse}
L.~Beers, M.~Weigand, H.~Nabizada, and A.~Fay, ``{MBSE Modeling Workflow for the Development of Automated Aircraft Production Systems},'' in \emph{28th International Conference on Emerging Technologies and Factory Automation (ETFA)}, 2023.

\bibitem{Katz.2018}
M.~Katz, S.~Sohrabi, H.~Samulowitz, and S.~Sievers, ``{Delfi: Online Planner Selection for Cost-Optimal Planning},'' in \emph{Ninth International Planning Competition}, 2018.

\bibitem{Muise.2022}
C.~Muise, F.~Pommerening, J.~Seipp, and M.~Katz, ``{PLANUTILS: Bringing Planning to the Masses},'' in \emph{32nd International Conference on Automated Planning and Scheduling (ICAPS).KA System Demonstrations}, 2022.

\bibitem{Weigand.2022}
M.~Weigand and A.~Fay, ``{Creating Virtual Knowledge Graphs from Software-Internal Data},'' in \emph{48th Annual Conference of the IEEE Industrial Electronics Society}, 2022.

\bibitem{Micheli2025UnifiedPlanning}
A.~Micheli, A.~Bit-Monnot, G.~Röger, E.~Scala, A.~Valentini, L.~Framba, A.~Rovetta, A.~Trapasso, L.~Bonassi, A.~E. Gerevini, L.~Iocchi, F.~Ingrand, U.~Köckemann, F.~Patrizi, A.~Saetti, I.~Serina, and S.~Stock, ``{Unified Planning: Modeling, manipulating and solving AI planning problems in Python},'' \emph{SoftwareX}, vol.~29, 2025.

\bibitem{ObjectManagementGroup.2024}
\BIBentryALTinterwordspacing
{Object Management Group}, ``{\textit{Systems Modeling Language (SysML), Version 2.0 beta 2}},'' 2024. [Online]. Available: \url{https://www.omg.org/spec/SysML/2.0/Beta2/}
\BIBentrySTDinterwordspacing

\end{thebibliography}
